\documentclass[runningheads]{llncs}
\usepackage[T1]{fontenc}
\usepackage{graphicx}
\usepackage{hyperref}
\hypersetup{colorlinks=true, linkcolor=blue, filecolor=blue, urlcolor=blue, citecolor=blue, hypertexnames=false}
\usepackage{color}

\usepackage{booktabs}
\usepackage{siunitx}
\usepackage{amsmath}
\usepackage{amsfonts}
\usepackage{amssymb}
\usepackage[boxed,lines numbered,no end,inoutnumbered]{algorithm2e}
\RestyleAlgo{ruled}            
\usepackage{multicol}
\usepackage{multirow}
\usepackage{array} 
\usepackage[caption=false]{subfig}

\usepackage{tabularx}
\usepackage{adjustbox}
\usepackage{float}

\makeatletter
\renewcommand\section{\@startsection{section}{1}{\z@}%
  {-7pt plus -1pt minus -1pt}%
  {0.8pt plus 0.4pt minus 0.4pt}%
  {\normalfont\large\bfseries}}

\renewcommand\subsection{\@startsection{subsection}{2}{\z@}%
  {-3pt plus -1pt minus -0.5pt}%
  {2pt plus 0.5pt minus 0.5pt}%
  {\normalfont\normalsize\bfseries}}

\renewcommand\subsubsection{\@startsection{subsubsection}{3}{\z@}%
  {-2pt plus -0.8pt minus -0.4pt}%
  {-0.35em}%
  {\normalfont\normalsize\bfseries}}
\makeatother

\let\oldthebibliography\thebibliography
\let\endoldthebibliography\endthebibliography
\renewenvironment{thebibliography}[1]{%
  \oldthebibliography{#1}%
  \setlength{\itemsep}{-0.6pt}%
  \setlength{\parskip}{0pt}%
}{\endoldthebibliography}

\begin{document}
\raggedbottom
\title{
Guiding Large Language Models with Genetic Programming-Evolved Heuristic Knowledge for Dynamic Multi-Mode Project Scheduling
}
\titlerunning{
Guiding LLM with GP-Evolved Heuristic Knowledge for DMRCPSP
}
%
\author{Yuan Tian\inst{1} \and Yi Mei\inst{1} \and
Mengjie Zhang\inst{1}}

\authorrunning{Y. Tian, et al.}

\institute{
Centre for Data Science and Artificial Intelligence \& \\School of Engineering and Computer Science, \\Victoria University of Wellington, PO Box 600, Wellington 6140, New Zealand\\
\email{\{yuan.tian, yi.mei, mengjie.zhang\}@ecs.vuw.ac.nz}}

\maketitle              
\begin{abstract}
In dynamic multi-mode project scheduling, activities have alternative execution modes and uncertain durations, while precedence relations and limited resources constrain their execution. Heuristic priority rules support fast online decisions, but their design requires substantial domain expertise.
Genetic programming (GP) hyper-heuristics can automatically evolve such rules. Large language models (LLMs), meanwhile, provide a flexible interface for interpreting scheduling information and explaining decisions. However, zero-shot LLM decisions may lack domain knowledge, consume many tokens, and vary across repeated queries. GP-evolved rules therefore provide a potential source of scheduling knowledge for guiding LLM decisions. Unlike existing LLM--GP hybrids that use LLMs to support heuristic evolution, we transfer knowledge in the reverse direction, using knowledge extracted from high-quality GP rules to guide an online LLM decision maker.
We extract knowledge from high-quality GP rules and inject it through Feature Selection, Feature Hint, Rule Reference, and Rule Follow. These mechanisms are evaluated in terms of scheduling performance, token consumption, decision stability, and the feature focus expressed in generated rationales. GP-derived guidance generally improves the unguided LLM, but its representation matters. Simplifying the decision context or supplying explicit decision logic is more effective than highlighting important features. Feature Selection offers the best token efficiency, whereas Rule Follow achieves strong performance at greater token cost. Guidance also improves decision stability and changes the features expressed in generated rationales.

\end{abstract}
\section{Introduction}

Project scheduling \cite{hartmann_updated_2022} is a fundamental problem in project management and planning. In dynamic multi-mode project scheduling, activities may have multiple execution modes, compete for limited resources, and have uncertain durations. Scheduling decisions, therefore, need to be made repeatedly as the project evolves, making fast and effective dispatching rules important.

Heuristic priority rules are widely used for dynamic scheduling because they can quickly select activities based on the current project state. However, manually designing effective rules requires domain expertise. Genetic programming (GP) hyper-heuristics \cite{zhang_survey_2023,tian_learning_2024} address this issue by automatically evolving priority rules from training instances. These rules combine scheduling features through mathematical expressions and can be applied to unseen instances, making them a useful source of learnt heuristic knowledge.

Recent work surveys the use of large language models (LLMs) in scheduling \cite{xiangSurveyLargeLanguage2026}. LLMs can combine scheduling information and heuristic knowledge at different levels of abstraction with the current context, and explain their decisions in natural language. However, an LLM does not inherently possess reliable scheduling knowledge or numerical rule-execution ability, so unguided decisions may be costly and unstable. Our goal is not to reproduce the numerical execution of a single GP rule, but to investigate whether knowledge extracted from GP--evolved rules at different levels of abstraction can guide an LLM's contextual scheduling decisions.

Existing LLM--GP research has largely transferred knowledge in the opposite direction, using LLMs to generate or improve heuristics within evolutionary search \cite{fangLeveragingLLMGenetic2025,xu_evospeak_2025}. This paper instead investigates GP-to-LLM transfer: GP evolution is completed first, and knowledge extracted from its high-quality rules then guides an LLM that selects one eligible activity--mode candidate at each decision point. We address the following research questions:
\begin{itemize}
    \item \textbf{RQ1:} Can GP-derived feature- and rule-level knowledge improve zero-shot LLM scheduling decisions, and which forms of guidance are most effective?
    \item \textbf{RQ2:} How do the guidance mechanisms affect token consumption, decision stability, and the feature focus expressed in generated rationales, and what effectiveness--efficiency trade-offs result?
\end{itemize}

To answer these questions, we extract feature- and rule-level knowledge from validated GP rules and evaluate four guidance mechanisms against an unguided LLM and the selected GP rule.

\section{Background and Related Work}
\subsection{Dynamic Multi-Mode Project Scheduling}
The dynamic multi-mode resource-constrained project scheduling problem (DMRCPSP) considers a project comprising a set of activities $J$ and renewable resources $R$. Each activity $i \in J$ is subject to precedence constraints and can be executed in one mode $m \in M_i$. The selected mode determines its resource demand $k_{i,m,r}$ for each resource $r \in R$ and its duration characteristics. An activity can start only after all its predecessors $j \in P_i$ have completed, and the total demand of concurrently executing activities cannot exceed the resource capacity $K_r$. For each activity-mode pair, the actual duration $d_{i,m}$ is uncertain and is revealed only when the activity starts, with its value sampled from a predefined distribution bounded by the optimistic and pessimistic durations $d^{\min}_{i,m}$ and $d^{\max}_{i,m}$. Scheduling decisions are therefore made dynamically as the project state evolves. The objective is to construct a feasible schedule satisfying precedence and resource constraints while minimising the project makespan.

\begin{figure}[t]
  \centering
  \begin{minipage}[b]{0.4\linewidth}
    \scriptsize
        \centering
        \renewcommand{\arraystretch}{0.6}
        \begin{tabular}{ccccccc}
    \toprule
    Act.$i$ & $P_i$&  Mode  & $\hat{d}_{i,m}$ & $d^{min}_{i,m}$  & $ d^{max}_{i,m}$ & $K_1$ \\
    \midrule
    1 & - & 1 & 0 & 0 & 0  & 0\\
    2 & 1 & 1 & 5 & 3 & 7  & 11\\
      & & 2 & 6 & 5 & 8  & 5\\
    3 & 1   & 1 & 4 & 3 & 7  & 7\\
      &     & 2 & 8 & 6 & 11 & 4\\
    4 & 2,3 & 1 & 4 & 3 & 6  & 8\\
      &     & 2 & 8 & 7 & 10 & 5\\
    5 & 2,4 & 1 & 4 & 2 & 5  & 4 \\
      &     & 2 & 7 & 4 & 8  & 4 \\
    6 & 4    & 1 & 3 & 2 & 5  & 9 \\
      &     & 2 & 9 & 7 & 10  & 6 \\ 
    7 & 5,6   & 1 & 0 & 0 & 0 &  0 \\
    \midrule
    \multicolumn{5}{l}{Resource availability} & & 10\\
    \bottomrule
\end{tabular}
    \newline    
    (a) Project information
  \end{minipage}
  \hspace{0.05\textwidth}
  \begin{minipage}[b]{0.45\linewidth}
    \centering
    \includegraphics[width=0.76\linewidth]{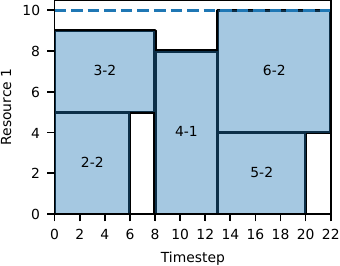}
    \newline
    (b) Example Schedule
  \end{minipage}
  \caption{An example project.}
  \label{fig:example_project}
\end{figure}

Figure \ref{fig:example_project} illustrates an example DMRCPSP instance. The project information is presented in Fig. \ref{fig:example_project} (a). The project consists of seven activities and one renewable resource, where activities 1 and 7 are the dummy start and end activities, respectively. Each non-dummy activity can be executed in one of two modes. Fig. \ref{fig:example_project} (b) shows an example schedule in terms of resource usage. Each rectangle represents a scheduled activity-mode pair and is labelled in the form \texttt{⟨act\_id-mode\_id⟩}. The width of a rectangle represents the activity duration, while its height indicates the corresponding resource demand.

\subsection{Genetic Programming Hyper-heuristics}
\begin{figure}[htbp]
  \centering
  \begin{minipage}[b]{0.5\linewidth}
    \centering
    \includegraphics[width=\linewidth]{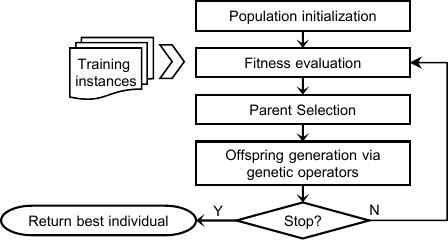}
    (a) GP evolution framework.
  \end{minipage}
  \hspace{0.05\textwidth}
  \begin{minipage}[b]{0.3\linewidth}
    \centering
    \includegraphics[width=0.5\linewidth]{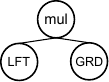}
    \newline
    (b) An example GP rule.
  \end{minipage}
  \caption{GP hyper-heuristic framework and priority rule representation.}
  \label{fig:GPHH_framework_representation}
\end{figure}

Genetic programming (GP) can be employed as a population-based hyper-heuristic to automatically evolve scheduling heuristics. As illustrated in Fig. \ref{fig:GPHH_framework_representation} (a), GP starts with a population of randomly generated individuals. Each individual is evaluated on a set of training instances, after which promising individuals are selected to generate offspring through genetic operators. This evolutionary process is repeated until a stopping criterion is reached, and the best evolved individual is returned.

GP individuals are commonly represented as tree-based mathematical expressions composed of scheduling features and arithmetic operators. In project scheduling, an evolved expression can serve as a priority function that assigns a priority value to each eligible activity-mode candidate. The candidate with the smallest value has the highest priority and is selected for scheduling. For example, the rule \texttt{LFT×GRD} represented in Fig. \ref{fig:GPHH_framework_representation} (b) evaluates the candidates shown in Table \ref{tab:priority_value_example}. Candidate (2,2) obtains the smallest priority value of 15×20=300 and is therefore selected. Through evolution, GP automatically searches for priority functions that produce effective scheduling decisions.

\begin{table}[t]
\centering
\caption{Example of priority value calculation and candidate selection.}
\label{tab:priority_value_example}
\vspace{-6pt}
\scriptsize
\begin{tabular}{lrrrrrrrrc}
\hline
\begin{tabular}[c]{@{}l@{}}Candidate\\(act., mode)\end{tabular}
& EST & EFT & LST & LFT & GRD & GRPW
& \begin{tabular}[c]{@{}r@{}}Priority\\value\end{tabular}
& Rank & Choice \\
\hline
$(2,1)$ & 0 & 10 & 0 & 10 & 40 & 30 & 400 & 4 & \\
$(2,2)$ & 0 & 15 & 0 & 15 & 20 & 30 & 300 & 1 & $\checkmark$ \\
$(3,1)$ & 0 &  8 & 2 & 10 & 32 & 15 & 320 & 2 & \\
$(4,2)$ & 0 & 12 & 3 & 15 & 24 & 13 & 360 & 3 & \\
\hline
\end{tabular}
\end{table}

\subsection{Large Language Model for Scheduling Decision Making}
LLMs have recently been explored as direct solvers for scheduling problems. End-to-end LLM-based approaches \cite{abgaryanStarjobDatasetLLMDriven2025,abgaryanLLMsCanSchedule2024} have been developed for job shop scheduling, where fine-tuned models directly generate scheduling solutions from problem instances. For dynamic flexible job shop scheduling, ReflecSched \cite{caoReflecSchedSolvingDynamic2026} employs heuristic-driven simulations to distil strategic experience that guides an LLM in making real-time scheduling decisions.
Another research direction uses LLMs to automatically design scheduling heuristics. EoH \cite{liuEvolutionHeuristicsEfficient2024} combines LLMs with evolutionary search for automatic heuristic generation. Population self-evolution \cite{huangAutomaticProgrammingLarge2026} has been introduced to design dispatching rules for dynamic job shop scheduling, while specialised LLM agents \cite{qiuEvoDREvolvingDispatching2026} have been used to generate and reflect on dispatching rules for dynamic flexible assembly flow shop scheduling. Evolutionary textual gradients \cite{huangAutomatedDesignScheduling2026} have also been employed to aggregate LLM-generated feedback and guide the iterative improvement of scheduling heuristics.

The symbolic structure of GP-evolved rules can make them more interpretable than black-box models \cite{meiExplainableArtificialIntelligence2022} and enable feature- and rule-level knowledge extraction. LLMs have been further integrated with GP to enhance heuristic evolution. LLM-generated individuals \cite{fangLeveragingLLMGenetic2025} have been introduced into GP populations to improve evolutionary search. More recently, knowledge extracted from high-quality GP heuristics has been used to warm-start evolution, support heuristic interpretation, and facilitate knowledge transfer across scheduling tasks \cite{xu_evospeak_2025}. These studies use the LLM primarily to create, improve, or interpret heuristics. In contrast, our framework reverses the usual LLM-to-GP knowledge flow: GP evolution is completed first, after which feature- and rule-level knowledge from evolved heuristics guides an LLM that directly makes online scheduling decisions. The framework also differs from ReflecSched, which distils strategic experience from heuristic-driven simulations, by transferring knowledge directly from a repository of symbolic GP rules. To the best of our knowledge, this GP-to-LLM transfer setting and LLM-based online decision making for the DMRCPSP have not been systematically investigated.

\section{Heuristic-Guided LLM Decision Making}
\subsection{LLM-Based Scheduling Decision Framework}

\begin{figure}[t]
    \centering
    \includegraphics[width=0.84\linewidth]{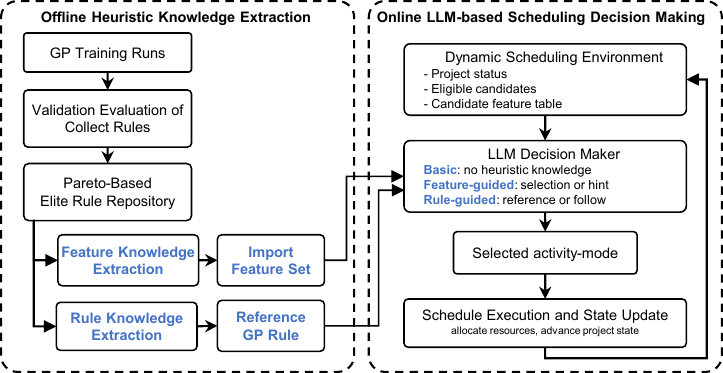}
    \caption{Overview of the heuristic-guided LLM decision-making framework for dynamic multi-mode project scheduling.}
    \label{fig:LLM_framework}
\end{figure}
We consider an online scheduling setting in which scheduling decisions are made dynamically as the project progresses. At each decision point, a heuristic-based decision mechanism is used to select an activity-mode candidate that can be executed immediately under the current project state. In the proposed framework, an LLM is used as the decision maker to perform this selection based on the information available at the current decision point.
As illustrated in Fig. \ref{fig:LLM_framework}, the dynamic scheduling environment first constructs a set of eligible activity-mode candidates. Let $S_t$ denote the project state at decision point $t$, and let $C_t=\{c_1, c_2, \ldots, c_{n_t}\}$ denote the corresponding candidate set. Each candidate $c_i=(a_i,m_i, \mathbf{x}_i)$ consists of an activity $a_i$, an execution mode $m_i$, and a scheduling feature vector $\mathbf{x}_i$. All candidates in $C_t$ satisfy the precedence and renewable resource requirements under the current project state and are therefore available for immediate execution. The current project status, feature definitions, and candidate feature table are provided to the LLM. Given optional heuristic knowledge $K$, the LLM selects one candidate according to $c^*_t=\mathcal{L}(S_t, C_t, K)$, where $\mathcal{L}$ denotes the LLM-based decision process. For the basic LLM, $K=\emptyset$, whereas the guided variants receive feature-level or rule-level heuristic knowledge extracted from evolved GP rules. The selected activity-mode is returned in a structured format and passed to the scheduling environment for execution. The selected candidate is started, the project simulation progresses, and the project state is updated according to the realised project dynamics. Once the next decision point is reached, a new set of immediately executable activity-mode candidates is constructed and provided to the LLM. This decision--execution--state update cycle continues until all project activities have been completed.

The prompt used at each decision point consists of a system instruction and a dynamically constructed user instruction. The system instruction specifies the scheduling decision task, candidate-selection constraints, decision procedure, explanation policy, and required output format. The user instruction provides the current project status, feature definitions, and eligible candidate table. Heuristic guidance is incorporated differently across the proposed LLM variants. As shown in Fig. \ref{fig:LLM_framework}, feature-guided variants use a selected feature set extracted from the GP rule repository, whereas rule-guided variants receive a selected GP priority rule. The detailed heuristic knowledge extraction and guidance mechanisms are introduced in the following subsections.

\subsection{Extraction of Heuristic Knowledge from GP Evolution}
\label{section:GP_evolution}
To obtain heuristic knowledge for guiding LLM decisions, we collect evolved GP rules from multiple independent GP runs. Rather than retaining only the final best individual from each run, the best individual of every generation is stored. This is relevant to the GP training process, where a seed rotation mechanism is used and the simulation seed changes across generations. Consequently, generation-best individuals are obtained under different simulated project conditions and may capture different scheduling behaviours. Retaining these individuals therefore provides a broader collection of evolved heuristics for subsequent knowledge extraction.
All collected rules are then evaluated on a separate validation set. GP evolution is performed using the training set, whereas the validation evaluation provides a consistent basis for comparing individuals collected from different runs and generations. For each rule, its effectiveness on the validation set and its rule size are recorded, representing scheduling quality and structural complexity, respectively.

Based on the trade-off between effectiveness and rule size, rules from the leading non-dominated fronts are prioritised to construct a Pareto-based elite rule repository. The corresponding knowledge extraction and guidance mechanisms are described in the following subsections.
\subsection{LLM Variants and Heuristic Guidance}
To investigate how evolved heuristic knowledge affects LLM-based scheduling decisions, we design five LLM variants with different levels and mechanisms of guidance. Table \ref{tab:llm_variants} summarises the differences between the five variants. The complete system and user prompt templates for all variants are provided in the supplementary material.

\begin{table}[t]
    \centering
    \caption{Summary of the LLM variants and heuristic guidance mechanisms.}
    \label{tab:llm_variants}
    \vspace{-6pt}
    \scriptsize
    \setlength{\tabcolsep}{8pt}
    \begin{tabular}{lcccc}
        \hline
        \textbf{Variant} & \textbf{Features} & \textbf{Feature Hint} & \textbf{GP Rule} & \textbf{Guidance} \\
        \hline
        Basic             & All      & No  & No  & None \\
        Feature Selection & Selected & No  & No  & Information restriction \\
        Feature Hint      & All      & Yes & No  & Attention guidance \\
        Rule Reference    & All      & No  & Yes & Advisory \\
        Rule Follow       & All      & No  & Yes & Prescriptive \\
        \hline
    \end{tabular}
\end{table}

\subsubsection{Basic:} The Basic variant receives no GP-derived heuristic knowledge. It is provided with the full feature definitions and candidate feature table and makes scheduling decisions based on the project information available at the current decision point. This variant serves as the baseline for evaluating the effect of heuristic guidance.

\subsubsection{Feature-level guidance:} To extract feature-level knowledge, a collection of high-quality rules is selected from the leading non-dominated fronts of the elite rule repository. The occurrence of each scheduling feature across these rules is analysed, where a feature is counted once for each rule in which it appears. Features are ranked by their occurrence frequency, and the most frequently occurring features covering 80\% of the cumulative feature occurrences are retained as the selected feature set. The selected features are incorporated into the LLM in two ways. \textbf{Feature Selection} restricts the feature definitions and candidate table to the selected feature set, thereby controlling the information available to the LLM. In contrast, \textbf{Feature Hint} retains all features but explicitly highlights the selected features as important decision cues. Thus, the former modifies information availability, whereas the latter provides attention guidance.

\subsubsection{Rule-level guidance:} For explicit rule guidance, a compact high-quality GP rule is selected from the elite rule repository. A compact rule is preferred to limit the reasoning effort and token consumption required to interpret and evaluate the priority expression across multiple candidates. The selected rule is supplied to both rule-guided variants. In \textbf{Rule Reference}, the GP rule is provided as advisory guidance: the LLM is encouraged to consider the rule when comparing candidates but may deviate when other candidate information provides stronger evidence. In \textbf{Rule Follow}, the rule acts as prescriptive guidance, and the LLM is instructed to evaluate the candidates according to the supplied priority expression and select the candidate with the best resulting priority.

\section{Experimental Study}
\subsection{Experimental Design}
The experiments investigate the scheduling performance and inference efficiency of LLM-based decision-making under different levels and mechanisms of heuristic knowledge guidance.
The experiments were conducted on dynamic multi-mode project scheduling instances with 30 activities and three execution modes per activity. Three precedence-constraint scenarios were considered, with order strengths \cite{demeulemeester_rangen_2003} of 0.75, 0.50, and 0.25. Five problem instances were used for each scenario. These scenarios represent different levels of scheduling decision complexity. A higher-order strength imposes stronger precedence constraints and generally results in fewer simultaneously eligible activity-mode candidates, whereas a lower-order strength produces less constrained and larger candidate sets. The scheduling features used for both GP training and LLM decision-making are provided in the supplementary material.

GP was trained 30 times separately for each precedence scenario, following the procedure described in Section \ref{section:GP_evolution}. The training, validation, and test sets share the same underlying project instances but use different realisations of actual activity durations. Heuristic knowledge extraction and rule selection were conducted based on the training and validation results only, without using the LLM test results. The GP configuration and evolutionary parameters are reported in the supplementary material.

Five LLM variants were evaluated to represent different levels and mechanisms of heuristic guidance. All variants used \texttt{deepseek-v4-flash} with thinking mode enabled, the reasoning effort set to \texttt{high}, and the temperature set to 0 to reduce sampling variability. The LLMs were used in a zero-shot decision setting. At each scheduling decision point, the model received only the information contained in the current prompt and had no access to previous decision contexts or conversation history. Unless explicitly modified by a variant, the system instructions, project-state representation, candidate constraints, and output format were kept unchanged.
For each precedence scenario, every LLM variant was evaluated on five test instances under the corresponding test duration realisations. Each instance was solved ten times using the same actual activity durations, allowing the variability of LLM-based decision-making to be observed under an identical scheduling environment.

\subsection{Experiment Analysis}

\begin{table}[t]
    \caption{Mean (standard deviation) normalised makespan obtained from 10 independent runs of the LLM variants under three scenarios.}
    \vspace{-6pt}
    \centering
    \scriptsize
    \renewcommand{\arraystretch}{0.86}
    \setlength{\tabcolsep}{2pt}
    \begin{adjustbox}{max width=\textwidth}
    \begin{tabular}{llllllll}
\hline
Scen. & Ins. & Basic      & Feature Hint             & Feature Selection                       & Rule Reference                                       & Rule Follow                                                        & GP Rule \\ \hline
0.75     & 1    & 1.72(0.07) & 1.73(0.09)($\approx$)    & 1.65(0.07)($\downarrow$)($\downarrow$) & 1.65(0.08)($\downarrow$)($\approx$)($\approx$)       & 1.58(0.00)($\downarrow$)($\downarrow$)($\downarrow$)($\downarrow$) & 1.58    \\
         & 2    & 1.39(0.08) & 1.50(0.14)($\uparrow$)   & 1.46(0.03)($\uparrow$)($\approx$)      & 1.39(0.12)($\approx$)($\downarrow$)($\approx$)       & 1.26(0.00)($\downarrow$)($\downarrow$)($\downarrow$)($\downarrow$) & 1.26    \\
         & 3    & 1.43(0.19) & 1.45(0.10)($\approx$)    & 1.44(0.03)($\approx$)($\approx$)       & 1.27(0.06)($\downarrow$)($\downarrow$)($\downarrow$) & 1.20(0.00)($\downarrow$)($\downarrow$)($\downarrow$)($\downarrow$) & 1.2     \\
         & 4    & 1.87(0.12) & 1.80(0.10)($\approx$)    & 1.82(0.05)($\approx$)($\approx$)       & 1.61(0.08)($\downarrow$)($\downarrow$)($\downarrow$) & 1.52(0.00)($\downarrow$)($\downarrow$)($\downarrow$)($\downarrow$) & 1.52    \\
         & 5    & 1.65(0.13) & 1.52(0.08)($\downarrow$) & 1.49(0.07)($\downarrow$)($\approx$)    & 1.42(0.08)($\downarrow$)($\downarrow$)($\downarrow$) & 1.34(0.00)($\downarrow$)($\downarrow$)($\downarrow$)($\downarrow$) & 1.34    \\
0.5      & 1    & 1.98(0.12) & 1.95(0.09)($\approx$)    & 1.88(0.11)($\downarrow$)($\downarrow$) & 1.73(0.08)($\downarrow$)($\downarrow$)($\downarrow$) & 1.78(0.07)($\downarrow$)($\downarrow$)($\downarrow$)($\uparrow$)   & 1.75    \\
         & 2    & 1.88(0.11) & 1.61(0.10)($\downarrow$) & 1.58(0.12)($\downarrow$)($\approx$)    & 1.70(0.17)($\downarrow$)($\approx$)($\approx$)       & 1.59(0.05)($\downarrow$)($\approx$)($\approx$)($\approx$)          & 1.59    \\
         & 3    & 2.42(0.16) & 2.10(0.08)($\downarrow$) & 2.16(0.13)($\downarrow$)($\approx$)    & 2.12(0.11)($\downarrow$)($\approx$)($\approx$)       & 2.03(0.10)($\downarrow$)($\approx$)($\approx$)($\approx$)          & 1.91    \\
         & 4    & 1.76(0.20) & 1.61(0.15)($\approx$)    & 1.64(0.10)($\approx$)($\approx$)       & 1.52(0.13)($\downarrow$)($\approx$)($\downarrow$)    & 1.56(0.16)($\downarrow$)($\approx$)($\approx$)($\approx$)          & 1.37    \\
         & 5    & 2.09(0.14) & 1.85(0.15)($\downarrow$) & 1.93(0.14)($\downarrow$)($\approx$)    & 1.84(0.12)($\downarrow$)($\approx$)($\approx$)       & 1.68(0.23)($\downarrow$)($\downarrow$)($\downarrow$)($\approx$)    & 1.59    \\
0.25     & 1    & 1.93(0.17) & 2.16(0.27)($\uparrow$)   & 1.57(0.14)($\downarrow$)($\downarrow$) & 1.88(0.19)($\approx$)($\downarrow$)($\uparrow$)      & 2.00(0.15)($\approx$)($\downarrow$)($\uparrow$)($\approx$)         & 2.0     \\
         & 2    & 2.04(0.08) & 2.19(0.14)($\uparrow$)   & 1.76(0.08)($\downarrow$)($\downarrow$) & 1.77(0.11)($\downarrow$)($\downarrow$)($\approx$)    & 1.70(0.07)($\downarrow$)($\downarrow$)($\approx$)($\approx$)       & 1.74    \\
         & 3    & 2.44(0.25) & 2.47(0.23)($\approx$)    & 2.00(0.14)($\downarrow$)($\downarrow$) & 1.86(0.10)($\downarrow$)($\downarrow$)($\downarrow$) & 2.01(0.05)($\downarrow$)($\downarrow$)($\approx$)($\uparrow$)      & 1.97    \\
         & 4    & 1.86(0.14) & 1.95(0.16)($\approx$)    & 1.54(0.09)($\downarrow$)($\downarrow$) & 1.75(0.15)($\approx$)($\downarrow$)($\uparrow$)      & 1.85(0.07)($\approx$)($\approx$)($\uparrow$)($\uparrow$)           & 1.91    \\
         & 5    & 2.64(0.20) & 2.99(0.28)($\uparrow$)   & 2.40(0.18)($\downarrow$)($\downarrow$) & 2.34(0.13)($\downarrow$)($\downarrow$)($\approx$)    & 2.26(0.12)($\downarrow$)($\downarrow$)($\approx$)($\approx$)       & 2.2     \\ \hline
\end{tabular}

    \end{adjustbox}
    \label{tab:test_performance}
\end{table}

\begin{table}[t]
    \caption{Counts of significant wins, ties, and losses of LLM variants compared with the LLM basic baseline.}
    \vspace{-6pt}
    \centering
    \scriptsize
    \renewcommand{\arraystretch}{0.86}
    \begin{tabular}{p{4.5cm} p{1.8cm} p{1.8cm} p{1.8cm}}
\toprule
Variant vs LLM Basic & Sig. Win & Sig. Tie & Sig. Loss \\
\midrule
Feature Selection & 11 & 3 & 1 \\
Feature Hint  & 4 & 7 & 4 \\
Rule Reference  & 12 & 3 & 0 \\
Rule Follow  & 13 & 2 & 0 \\
\bottomrule
\end{tabular}

    \label{tab:win_tie_loss}
\end{table}
\subsubsection{Scheduling Performance.}
Table~\ref{tab:test_performance} reports the mean and standard deviation of the normalised makespan over ten repeated runs. The GP rule is included as a heuristic baseline. The significance markers are obtained using pairwise Wilcoxon signed-rank tests against the preceding columns in the same row. An upward arrow ($\uparrow$) indicates a significantly larger normalised makespan, a downward arrow ($\downarrow$) indicates a significantly smaller normalised makespan, and a tilde ($\approx$) indicates no significant difference.
Overall, most forms of heuristic guidance improve the Basic LLM. As summarised in Table~\ref{tab:win_tie_loss}, Feature Selection, Rule Reference, and Rule Follow obtain more significant wins than losses against the Basic variant, whereas Feature Hint shows less consistent improvement. Rule Follow performs best in the high-order-strength scenario, but in the medium- and low-order-strength scenarios its advantage becomes less clear, with Rule Reference and Feature Selection often achieving comparable results. This suggests that explicit rule-level guidance is useful but does not uniformly dominate. In some instances, Feature Selection even outperforms both the GP rule and Rule Follow, indicating that scenario-specific feature filtering can sometimes provide a more robust form of guidance than enforcing a single selected priority rule.
Although the GP rule is generally a strong baseline, it is not uniformly superior across all test instances. This also suggests that the selected rule may have some instance-dependent behaviour under unseen activity-duration realisations, which partly explains why Rule Follow does not always achieve the best performance.

    
\subsubsection{Token Consumption and Composition.}
\begin{figure}[t]
    \centering
    \includegraphics[width=0.76\linewidth]{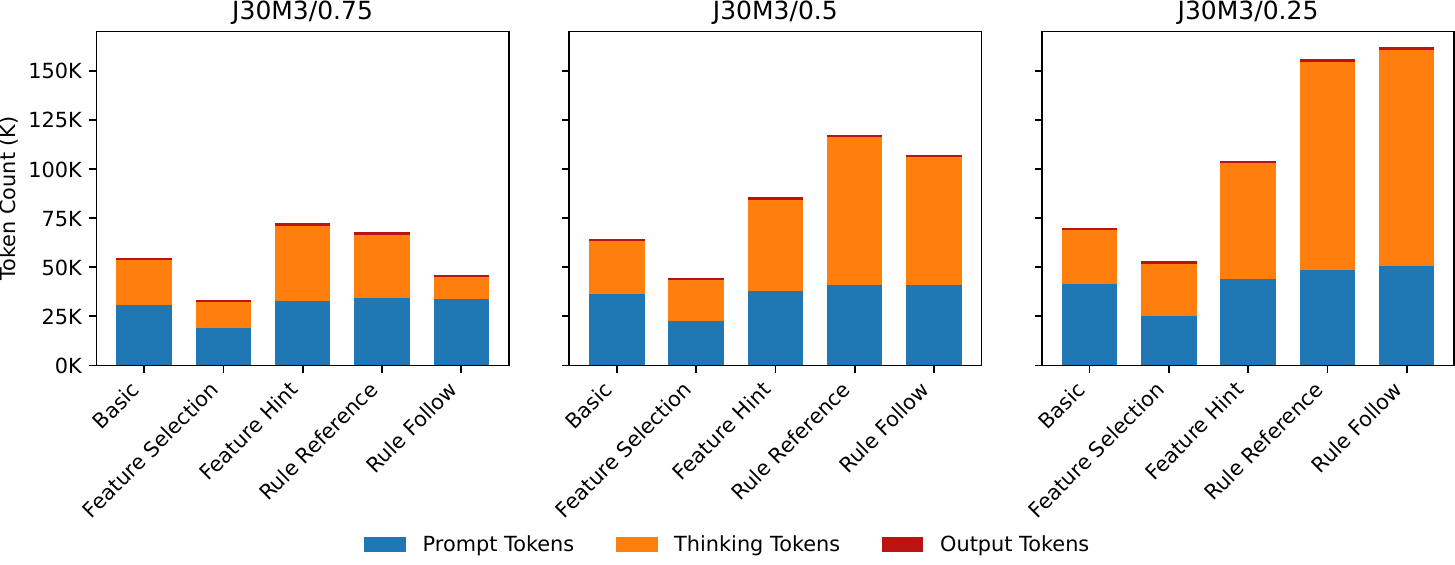}
    \caption{Token consumption across LLM variants and scenarios.}
    \label{fig:token_composition}
\end{figure}

\begin{figure}[!htbp]
    \centering
    \includegraphics[width=0.4\linewidth]{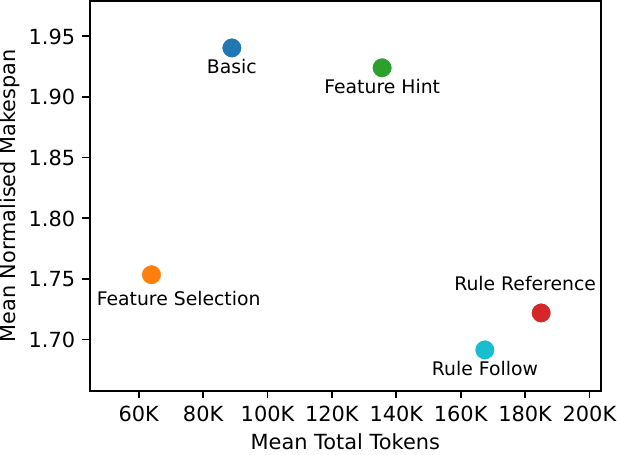}
    \caption{Token--makespan trade-off across different LLM-based scheduling strategies.}
    \label{fig:token_performance_tradeoff}
\end{figure}
Figure~\ref{fig:token_composition} reports the token consumption of each LLM variant, aggregated by order-strength scenario and separated into input prompt tokens, thinking tokens, and output tokens. Overall, token consumption increases as the order strength decreases. This is because lower order strength imposes fewer precedence constraints, allowing more activities to become eligible at the same decision point. As a result, the candidate table becomes larger and the LLM needs to compare more activity-mode alternatives, leading to both longer prompts and more thinking tokens.
Across all scenarios, most tokens are spent on thinking rather than on the final response. This is especially evident for Rule Reference and Rule Follow, where the LLM needs to interpret the supplied priority rule and compare candidates according to rule-related calculations. Rule Reference consumes even more thinking tokens than Rule Follow in some scenarios, possibly because the rule is used as soft guidance, requiring the model to consider both the rule and other candidate information.
The input prompt tokens mainly correspond to the prompt content, including project status, feature definitions, and the candidate information table. Feature Selection requires the fewest prompt tokens because only the selected scenario-specific features are included, whereas the other variants receive the full feature set and therefore have similar input-token costs. Output tokens are small across all variants, since the model only returns the selected activity-mode pair and a brief reason.

Figure~\ref{fig:token_performance_tradeoff} shows the performance--token trade-off. Feature Selection provides the best efficiency, achieving improved makespan with the lowest token consumption. Rule Follow achieves strong performance but requires substantially more tokens due to explicit rule evaluation. The remaining variants are less attractive: Feature Hint increases token usage without clear performance gains, while Rule Reference consumes the most tokens but performs similarly to Rule Follow. Therefore, Feature Selection is preferable when token efficiency is important, whereas Rule Follow is a better choice when scheduling performance is the main priority.

\subsubsection{Feature Mentions in LLM Rationales.}
\begin{figure}[!htbp]
    \centering
    \includegraphics[width=0.75\linewidth]{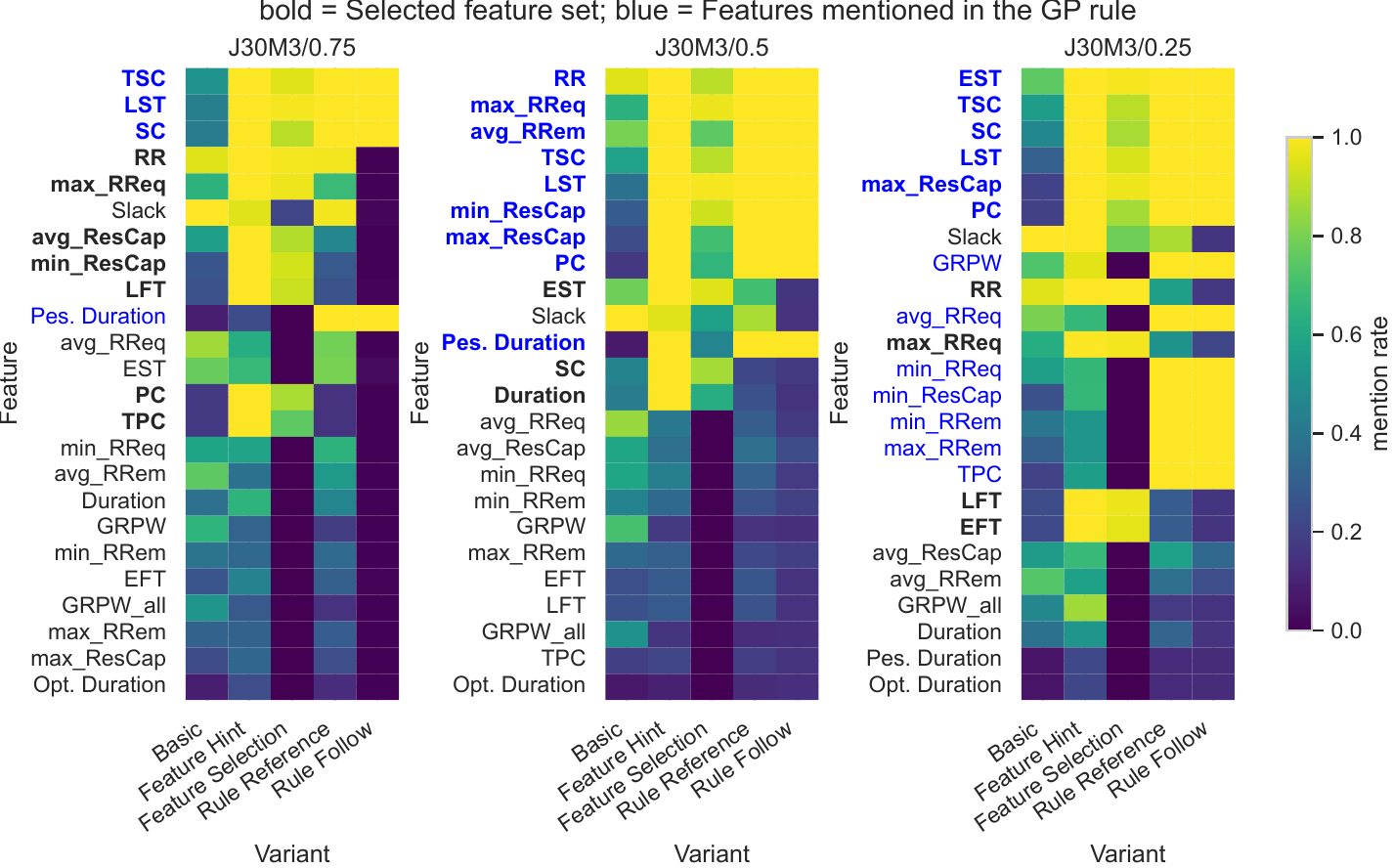}
    \caption{Feature mention rates in LLM-generated rationales across three order-strength scenarios.}
    \label{fig:feature_mention_rate}
\end{figure}
We examine whether heuristic guidance changes the scheduling features explicitly referred to in the LLM-generated decision reasons. For each variant and order-strength scenario, the reasoning content and returned reason were semantically matched against the scheduling feature definitions; therefore, a detected mention does not necessarily mean that the exact feature name appeared verbatim. Figure~\ref{fig:feature_mention_rate} shows the proportion of decisions in which each feature was identified. The results show that guidance changes the feature focus expressed by the LLM. Feature Selection concentrates mentions on the retained features. Interestingly, \texttt{Slack} remains frequently identified even though it is not included in the selected feature set, suggesting that the model can reconstruct the notion of scheduling flexibility from related temporal information (e.g., \texttt{LFT-EFT}) rather than relying only on explicitly named inputs. Feature Hint increases references to highlighted features without consistently improving performance, while the rule-guided variants shift mentions towards GP-rule features. Overall, GP-derived guidance affects both scheduling decisions and the rationales used to articulate them.

\subsubsection{Stability of LLM Decisions.}
\begin{figure}[!htbp]
    \centering
    \includegraphics[width=0.72\linewidth]{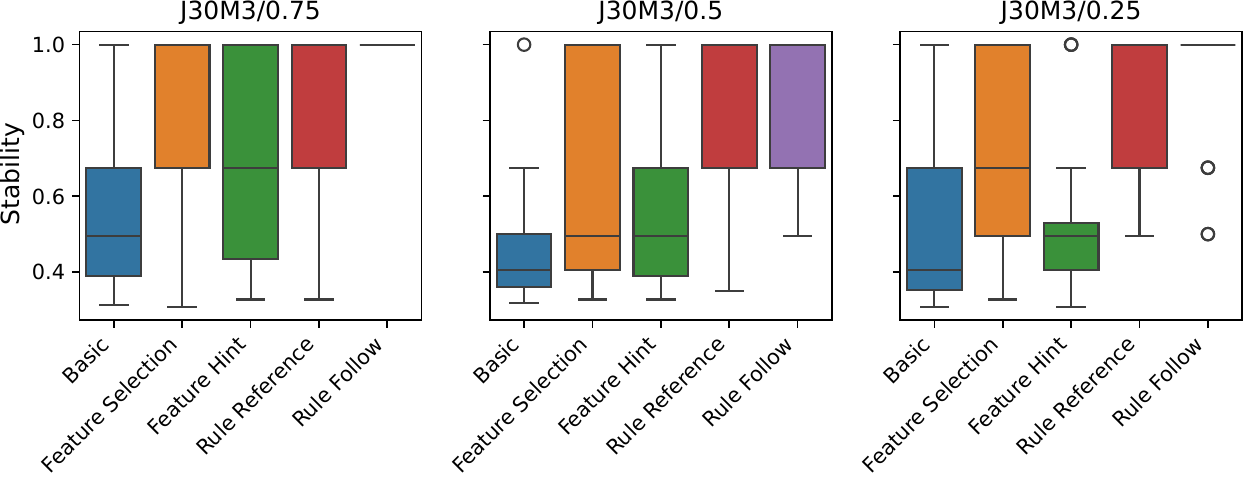}
    \caption{Decision stability of LLM variants across sampled decision situations.}
    \label{fig:decision_stability}
\end{figure}

Figure~\ref{fig:decision_stability} analyses the stability of LLM decisions on sampled decision situations. This analysis is separate from the full scheduling test runs. For each order-strength scenario, 25 decision situations were sampled from the solved instances. Each LLM variant was queried ten times on each situation, and the selected activity-mode pairs were recorded. We computed the normalised entropy \(H\) of the selection distribution and used \(1-H\) as the decision stability score. A larger value therefore indicates that the LLM more consistently selects the same candidate under repeated queries.

The Basic variant shows relatively low stability, suggesting that unguided LLM decisions can vary even when the prompt and decision situation are unchanged. Feature Selection slightly improves stability, likely because the reduced candidate table removes some distracting information. Feature Hint and Rule Reference achieve higher stability because the provided guidance constrains the model's decision focus. Rule Follow is the most stable in the high-order-strength scenario, but its stability decreases in the medium- and low-order-strength scenarios. This indicates that instructing the LLM to follow a rule does not fully eliminate decision variability, possibly due to rule-interpretation or calculation errors when the candidate set becomes larger.

\subsubsection{Generated Decision Explanations.}
\begin{table}[t]
\centering
\caption{Representative decisions and natural-language reasons produced by different LLM variants for the same decision situation in J30M3/0.5.}
\label{tab:reason_example}
\vspace{-6pt}
\scriptsize
\renewcommand{\arraystretch}{0.9}
\setlength{\tabcolsep}{2pt}

\begin{tabularx}{\linewidth}{p{0.20\linewidth} p{0.12\linewidth} X}
\hline
\textbf{Variant} & \textbf{Choice} & \textbf{Generated explanation} \\
\hline
Basic
& $(19,1)$
& Critical activity (slack=0) with shortest duration (6) to expedite project completion. \\

Feature Hint
& $(7,2)$
& Highest TSC (15) and critical (LS=0) with moderate duration (18) and low max\_RReq (5). \\

Feature Selection
& $(7,2)$
& Selected due to criticality (LS=0), highest number of successors (TSC=15), and moderate duration and resource consumption.\\

Rule Reference
& $(7,2)$
& Lowest priority rule value (4.45) due to small \texttt{LS} and favourable resource trade-off.\\

Rule Follow
& $(7, 2)$
& Selected candidate (7,2) with lowest priority rule value of 4.4509.\\
\hline
\end{tabularx}
\end{table}

Table~\ref{tab:reason_example} presents representative choices and one-sentence reasons generated for the same decision situation. Basic selects candidate $(19,1)$ based mainly on zero slack and the shortest duration. In contrast, the feature- and rule-guided variants select $(7,2)$ and express a broader decision focus, jointly considering activity criticality, duration, and resource consumption. This example qualitatively illustrates how heuristic guidance can shape both candidate selection and the rationale expressed by the LLM.

\section{Conclusions and Future Work}

This paper introduced a GP-to-LLM knowledge-transfer framework for online dynamic multi-mode project scheduling. Its main contributions are feature- and rule-level mechanisms that guide LLM decisions with knowledge extracted from evolved priority rules, an empirical comparison of four guidance variants with an unguided LLM, and analyses of scheduling performance, token usage, decision stability, and generated rationales. Regarding RQ1, Feature Selection, Rule Reference, and Rule Follow generally outperform the unguided Basic variant, whereas Feature Hint is less reliable. Regarding RQ2, Feature Selection offers the best token efficiency, while Rule Follow achieves strong performance at greater token cost. Guidance can also reduce decision variability and shift the feature focus expressed in generated rationales. Overall, the results demonstrate that evolved heuristic knowledge can guide an LLM at multiple levels of abstraction, with distinct performance--efficiency trade-offs.
Future work will select rules dynamically from a rule repository according to the current decision situation, apply multi-stage candidate filtering to support larger project instances, and extract additional forms of GP knowledge.

%
%
\bibliographystyle{splncs04}
\bibliography{references}
\end{document}